\documentclass{article}
\usepackage{comment}
\usepackage{float}
\usepackage[accepted]{icml2018}
\usepackage{amsmath,amssymb}

\usepackage{microtype}
\usepackage{graphicx}
\usepackage{subfigure}
\usepackage{booktabs} 

\usepackage[table]{xcolor}
\usepackage{epstopdf}
\usepackage{amsmath}
\usepackage{tcolorbox}
\usepackage{array}
\usepackage{hyperref}
\usepackage{amsmath,amssymb}
\usepackage{tikz}
\usepackage{float}

\newtheorem{theorem}{Theorem}[section]

\newtheorem{lemma}[theorem]{Lemma}
\newtheorem{remark}[theorem]{Remark}

\usepackage[utf8]{inputenc}
\usepackage[english]{babel}
\usepackage{natbib}

\usepackage[utf8]{inputenc}
\usepackage[english]{babel}
\usepackage{fancyhdr}
\usepackage{lastpage}
 
\begin{document}

\onecolumn
\icmltitle{WNGrad: Learn the Learning Rate in Gradient Descent}
\icmlsetsymbol{equal}{*}
\begin{icmlauthorlist}
\icmlauthor{Xiaoxia Wu }{equal,to,goo}
\icmlauthor{Rachel Ward }{equal,to,goo}
\icmlauthor{L{\'e}on Bottou }{goo}

\end{icmlauthorlist}

\icmlaffiliation{to}{Department of Mathematics, University of Texas at Austin, Austin, TX}
\icmlaffiliation{goo}{Facebook AI Research, New York, New York, USA}

\icmlcorrespondingauthor{Xiaoxia Wu}{xwu@math.utexas.edu}

\icmlkeywords{Stochastic Gradient Descent, Learning Rate, Reparametrization}

\vskip 0.3in


%
\begin{abstract}
Adjusting the learning rate schedule in stochastic gradient methods is an important unresolved problem which requires tuning in practice.  If certain parameters of the loss function such as smoothness or strong convexity constants are known, theoretical learning rate schedules can be applied.  However, in practice, such parameters are not known, and the loss function of interest is not convex in any case. The recently proposed \emph{batch normalization} reparametrization is widely adopted in most neural network architectures today because, among other advantages,  it is robust to the choice of Lipschitz constant of the gradient in loss function, allowing one to set a large learning rate without worry.  Inspired by batch normalization, we propose a general nonlinear update rule for the learning rate in batch and stochastic gradient descent so that the learning rate can be initialized at a high value, and is subsequently decreased according to gradient observations along the way.  The proposed method is shown to achieve robustness to the relationship between the learning rate and the Lipschitz constant, and near-optimal convergence rates in both the batch and stochastic settings ($O(1/T)$ for smooth loss in the batch setting, and $O(1/\sqrt{T})$ for convex loss in the stochastic setting) .  We also show through numerical evidence that such robustness of the proposed method extends to highly nonconvex and possibly non-smooth loss function in deep learning problems. 
 Our analysis establishes some first theoretical understanding into the observed robustness for batch normalization and weight normalization.  \let\thefootnote\relax\footnote{$*$ Equal contribution. {\footnotesize $1$} Department of Mathematics, University of Texas at Austin, Austin, TX.  {\footnotesize $2$} Facebook AI Research, New York, New York, USA.  Note that the paper by \cite{ward2019adagrad} has greatly  subsumes this manuscript.}
\end{abstract}
\section{Introduction}
\label{introduction}

Recall the standard set-up for gradient descent: we consider the general problem of minimizing a ``loss function" $f: \mathbb{R}^d \rightarrow \mathbb{R}$,
\begin{equation}
\label{gd:general}
\min_x \quad f(x)
\end{equation}
and given access only to first-order/gradient evaluations of $f$, we iteratively move in the direction of the negative gradient until convergence: $x_{j+1} \leftarrow x_j - \eta_j \nabla f(x_j)$.  Gradient descent enjoys nice convergence guarantees if the learning rate  $\eta_j = \eta$ is tuned just right according to the scale of the smoothness of the gradient function $\nabla f(x)$; on the other hand, if the learning rate is chosen slightly larger than the optimal value, gradient descent with constant learning rate can oscillate or even diverge.  Thus, in practice, one instead uses iteration-dependent learning rate $\eta_j,$ chosen via line search methods 
 \cite{bertsekas1999nonlinear,Nocedal2006}.
Line search methods work well in the ``batch" set-up where the gradients $\nabla f(x_j)$ are observed \emph{exactly}, but notoriously become less effective in the \emph{stochastic} setting, where only noisy gradient evaluations are given.   Recall the standard setting for stochastic gradient descent: Instead of observing a full gradient $\nabla f(x_k)$ at iteration $k$, we observe a stochastic gradient $g_k$, or a random vector satisfying $\mathbb{E}(g_k) = \nabla f(x_k)$ and having bounded variance $\mathbb{E} \| g_k \|^2 \leq G^2$.  Stochastic gradient descent is the optimization algorithm of choice in deep learning problems, and, more generally, in many large-scale optimization problems where the objective function $f$ can be expressed as a sum of a number component functions $f_i$, of which only have access to a subset (the so-called ``training data").  

In the stochastic setting, the issue of how to choose the learning rate is less resolved.  There are different guidelines for setting the learning ``schedule" $\eta_1, \eta_2, \dots$, each guideline having its own justification in the form of a convergence result given a set of structural assumptions on the loss function $f$.  The classical Robbins/Monro theory \cite{rm51} says that if the learning rate is chosen such that
\begin{equation}
\label{step:rm}
\sum_{k=1}^{\infty} \eta_k = \infty \quad \text{and} \quad \sum_{k=1}^{\infty} \eta_k^2 < \infty,
\end{equation}
and if the loss function is sufficiently smooth, then $\lim_{k \rightarrow \infty} \mathbb{E}[ \|\nabla f(x_k)\|^2 ] = 0$  (\cite{bcn16}, Corollary 4.12).   
If the loss function is moreover strongly convex, the stochastic gradient update 
$x_{k+1} \leftarrow x_k - \eta_k g_k$ will converge in expectation to the minimizer .  

If the loss function is convex but not necessarily smooth, then setting $\eta_k = c/\sqrt{k}$ results in a convergence guarantee of the form $\mathbb{E}[ f(x_k) - f^{*} ] \leq O(\log(k)/\sqrt{k})$, with an optimal constant if $c$ is chosen properly to depend on the stochastic variance $G^2$ (\cite{shamir2013stochastic}, Theorem 2). 
If the loss function  is $\mu$-\emph{strongly} convex and has $L$-Lipschitz smooth gradient, then setting $\eta_k = c/k$ where $c$ is sufficiently small compared to $\frac{\mu}{L}$ gives $\mathbb{E}[ f(x_k) - f^{*} ] \leq O(1/k)$ (\cite{bcn16}, Theorem 4.7). 

If, moreover, the loss function can be expressed as the average of a number of component functions $f_i$, each of which is itself convex, and if the noisy gradient direction at each iteration is actually the direction of the exact gradient of one of the component functions chosen i.i.d. uniformly from the universe of component functions, and if a bound on the ``consistency" parameter $\sigma^2 = \mathbb{E}_i[ \| f_i(x^{*}) \|^2 ]$ is known,  then one may take a constant learning rate $\eta_k = \eta$ chosen in just the right way with respect to $\mu$, $L,$ and $\sigma^2$ to achieve \emph{exponential} convergence, in expectation, up to a radius around  the optimal solution \cite{bm11, nsw14}.

Thus, in the stochastic setting, there is no clear ``best choice" for the learning rate.  In many deep learning problems, where the underlying loss function is highly non-convex, one often tests several different learning rate schedules of the form 
$$
\eta_k = \frac{\eta_0}{1 +k/\tau}, \text{ or } \eta_k = \frac{\eta_0}{1 + \sqrt{k}/\tau}, \text{ or } \eta_k = \frac{D}{kG};
$$
where $D$ is the maximal diameter of the feasible set, and $G$ is the norm of the current gradient or an average of recent gradients; the schedule which works best on the problem at hand is then chosen. Another popular and effective choice is to start with a constant learning rate $\eta_0$ which gives good empirical convergence results or start with a small one followed by a warmup scheme \cite{GoyalDGNWKTJH17}, maintain this constant learning rate for a fixed number of epochs over the training data, then decrease the learning rate $\eta_1 \leftarrow 0.1\eta_0$, and repeat this process until convergence  . 
\subsection{Adaptive Learning Rate Rules}
In the stochastic setting, it can be advantageous to set different learning rates for different component functions $f_i$ (or for different coordinates), with larger learning rates for components with smaller gradients, and smaller learning rates for components with larger gradients, to balance their respective influences. This heuristic is theoretically justified in some cases. The family of adaptive gradient (AdaGrad) algorithms \cite{duchi2011adaptive} dynamically update each coordinate learning rate by the reciprocal of the root-mean-square of the elements of the gradients for that coordinate which have been observed so far.  AdaGrad has rigorous theoretical backing: it provably achieves the optimal $\mathbb{E}[ f(x_k) - f^{*} ] \leq O(1/\sqrt{k})$ regret guarantee in the convex setting, with a better constant compared to plain stochastic gradient descent depending on the geometry of the problem.  Despite originally being designed in the convex setting, AdaGrad has proven to be very useful beyond the convex set-up -- in particular, it improves convergence performance over standard stochastic gradient descent in settings where data is sparse and sparse parameters are more informative; such examples abound in natural language processing.  Several subsequent modifications to  AdaGrad have been proposed to combat this accumulation including Adadelta \cite{zeiler2012adadelta}, RMSprop \cite{hinton2012neural}, Adam \cite{kingma2014adam} and AdaBatch  \cite{defossez2017adabatch}; however, these algorithms (except AdaBatch) come with no guarantees of convergence.  These adaptive subgradient methods cannot be applied as a general panacea for the learning rate problem, however as they result in biased gradient updates which change the underlying optimization problem.  The recent paper \cite{wilson2017marginal} provides evidence that while these methods do speed up training time in neural network applications, they nevertheless result in worse generalization error compared to simple methods such as plain stochastic gradient descent with a single learning rate. 

Another line of work on adaptive learning rates \cite{nsw14, zhao2015stochastic} consider importance sampling in stochastic gradient descent in the setting where the loss function can be expressed as a sum of component functions, and provide precise ways for setting different constant learning rates for different component functions based on their Lipschitz constants; if the sampling distribution over the parameters is weighted so that parameters with smaller Lipschitz constants are sampled less frequently, then this reparametrization affords a faster convergence rate, depending on the \emph{average} Lipschitz constant between all parameters, rather than the \emph{largest} Lipschitz constant between them.  Of course, in practice, the Lipschitz constants are not known in advance, and must be learned along the way.  

This begs the question: if we take a step back to the batch/non-stochastic gradient descent setting, is it possible to learn even a \emph{single} Lipschitz constant, corresponding to the gradient function $\nabla f$, so that we can match the convergence rate of gradient descent with optimized constant learning rate which requires knowledge of the Lipschitz constant beforehand? To our knowledge, this question has not been addressed until now.

\subsection{Weight Normalization}
To answer this question, we turn to simple reparametrizations of weight vectors in neural networks which have been proposed in recent years and have already gained widespread adaptations in practice due to their effectiveness in accelerating training times without compromising generalization performance, while simultaneously being robust to the tuning of learning rates.   The celebrated batch normalization \cite{ioffe2015batch} accomplishes these objectives by normalizing the means and variances of minibatches in a particular way which reduces the dependence of gradients on the scale of the parameters or their initial values, allowing the use of much higher learning rates without the risk of divergence.  Inspired by batch normalization, the \emph{weight normalization} algorithm  \cite{salimans2016weight} was introduced as an even simpler reparametrization, also effective in making the resulting stochastic gradient descent more robust to specified learning rates and initialization.  The weight normalization algorithm, roughly speaking, reparametrizes the loss function in polar coordinates, and runs (stochastic) gradient descent with respect to polar coordinates: If the loss function is $f(x)$ where $x$ is a $d$-dimensional vector, then weight normalization considers instead  $x = \frac{r}{\| v \|} v,$ where $v$ is a $d$-dimensional vector, $r$ is a scalar, and $\| v \|$ is the Euclidean norm of $v$.  The analog of the weight normalization algorithm in the batch gradient setting would simply be gradient descent in polar coordinates as follows:  
\begin{align}
\label{wnupdate}
v_{k+1} &= v_k - \eta \nabla_v f(\frac{r_k}{\| v_k \|} v_k ) \nonumber \\
&= v_k - \eta \frac{r_k}{\| v_k \|}P_{v_k^{\perp}}( \nabla f(\frac{r_k}{\| v_k \|} v_k )); \nonumber \\
r_{k+1} 
&= r_k - \eta \nabla_r f(\frac{r_k}{\| v_k \|} v_k ) \nonumber\\
&= r_k -\eta  \langle{ \nabla f(\frac{r_k}{\| v_k \|} v_k), \frac{v_k}{\| v_k \|} \rangle}  
\end{align}
where $P_{v^{\perp}}(u)$ denotes the orthogonal projection of $u$ onto the subspace of co-dimension orthogonal to $v$.
One important feature of note is that, since the gradient of $f$ with respect to $v$ is orthogonal to the current direction $v$, the norm $\| v_k \|$ grows monotonically with the update, thus effectively producing a dynamically-updated decay in the \emph{effective} learning rate $\frac{r \eta}{\| v \|^2}$.  More precisely, considering weight normalization in the batch setting, restricted to the unit sphere (fixing $r_k = 1$), the gradient update reduces to 
\begin{align}
\label{wnupdate_sphere}
\frac{v_{k+1}}{\| v_k \|} &= \frac{v_k}{\| v_k \|} - \frac{\eta}{\| v_k \|^2}P_{v_k^{\perp}}(\nabla f(\frac{v_k}{\| v_k \|})); \nonumber \\
\| v_{k+1} \|^2 &= \| v_k \|^2 + \frac{\eta^2}{\| v_k \|^2} \| P_{v_k^{\perp}}(\nabla f(\frac{v_k}{\| v_k \|})) \|^2 
\end{align}
\subsection{Our contributions}
Weight normalization (and, to an even larger extent, batch normalization) has proven in practice to be very robust to the choice of the scale of Lipschitz constant $\eta$.  Inspired by this, and in a first attempt at theoretical understanding of such normalization, we are inspired to consider the following method for updating the learning rate in batch and stochastic gradient descent more generally: starting from $x_1 \in \mathbb{R}^d$ and $b_1 > 0,$ repeat until convergence
\begin{align}
\label{WNGrad}
x_{k+1} &= x_k - \frac{1}{b_k} \nabla f(x_k); \nonumber \\
b_{k+1} &= b_k + \frac{1}{b_k}  \| \nabla f(x_k) \|^2
\end{align}
As a nod to its inspiration, weight normalization, we call this algorithm WNGrad, but note that the update can also be interpreted as a close variant of AdaGrad with the dynamic update applied to a single learning rate; indeed, WNGrad $b$-update satisfies
\begin{align}
b_{k+1}^2 &= b_k^2 + 2\| \nabla f(x_k) \|^2 +  \frac{1}{b_k^2}  \|  \nabla f(x_k) \|^4 \nonumber \\
&= b_k^2 + 2\|  \nabla  f(x_k) \|^2 + O( \|  \nabla f(x_k) \|^4)  \nonumber 
\end{align}
which matches the coordinate-wise update rule in AdaGrad if $\nabla f$ is one dimension.
Nevertheless, WNGrad update \eqref{WNGrad} offers some insight and advantages over the family or modifications/improvements of AdaGrad update -- first, it gives a precise correspondence between the accumulated gradient and current gradient  in the update of the $b_k$.  Additionally, it does not require any square root computations, thus making the update more efficient.  

In this paper, we provide some basic theoretical guarantees about WNGrad update. Surprisingly, we are able to provide guarantees \emph{for the same learning rate update rule} in both the batch and stochastic settings.

In the batch gradient descent setting, we show that WNGrad will converge to a weight vector $x_T$ satisfying $\| \nabla f(x_T) \|^2 \leq \epsilon$ in at most  $T = O(\frac{\left(f(x_1)-f^*+L\right)^2}{\epsilon})$ iterations,  if $f$ has $L$-Lipschitz smooth gradient.  The proof involves showing that if $b_k$ grows up to the critical level $b_k \geq L,$ it \emph{automatically stabilizes}, satisfying $b_k \leq CL$ for all time\footnote{C is a constant factor}.  This should be compared to the standard gradient descent convergence rate using constant learning rate $\eta,$ which in the ideal case $\eta = 1/L$ achieves $O(\frac{L}{\epsilon})$ convergence rate, but which \emph{is not guaranteed to converge at all} if the learning rate is even slightly too big, $\eta \geq 2/L.$  Thus, WNGrad is a provably robust variant to gradient descent which is provably robust to the scale of Lipschitz constant, when parameters like the Lipschitz smoothness are not known in advance.

On the other hand, in the stochastic setting, the $b_k$ update in WNGrad has dramatically different behavior, growing like $O(\frac{\sqrt{k}}{G}),$ where $G$ is a bound on the variance of the stochastic gradients.  As a result, in the stochastic setting, we also show that WNGrad,  achieves the optimal $O(1/\sqrt{T})$ rate of convergence for convex loss functions, and moreover settles in expectation on the ``correct" constant, $b_k \sim \frac{\sqrt{k}}{G}$.  Thus, WNGrad also works robustly in the stochastic setting, and finds a good learning rate. 

We supplement all of our theorems with numerical experiments, which show that WNGrad competes favorably to plain stochastic gradient descent in terms of  robustness to the Lipschitz constant of the loss function, speed of convergence, and generalization error, in training neural networks on two standard data sets.  
\section{WNGrad for Batch Gradient Descent}

Consider a smooth function $f: \mathbb{R}^d \rightarrow \mathbb{R}$ with $L$-Lipschitz continuous gradient (denoted $f \in C_L^1$): for any $x, y \in \mathbb{R}^d$,
$$
\| \nabla f(x) - \nabla f(y) \| \leq L \| x - y \|
$$
and the optimization problem
\begin{equation}
\nonumber
\min_x \quad f(x).
\end{equation}

With knowledge of the Lipschitz constant $L$, the standard gradient descent update with constant learning rate iterates, starting at $x_1 \in \mathbb{R}^d$,
\begin{equation}
\label{grad_descend}
x_j \leftarrow x_{j-1} - \eta \nabla f(x_{j-1}).
\end{equation}

The following convergence result is classical (\cite{nesterov1998introductory}, $(1.2.13)$).
\begin{lemma}
\label{lem:classical}
Suppose that $f \in C_L^1$ and that $f^{*} > -\infty$.  
Consider gradient descent with constant learning rate $\eta > 0$.

If $\eta = \frac{\delta}{L}$ and $\delta \leq 1$, then
$$
\min_{j=1:T} \| \nabla f(x_j) \|^2 \leq \epsilon
$$
after at most a number of steps 
$$
T =\frac{2L(f(x_1)-f^{*})}{\delta \epsilon};
$$
On the other hand, gradient descent can oscillate or diverge once $\eta \geq \frac{2}{L}$.
\end{lemma}

Note that this result requires the knowledge of Lipschitz constant $L$ or an upper bound estimate.  Even if such a bound is known, the algorithm is quite conservative; the Lipschitz constant represents the \emph{worst case} oscillation of the function $\nabla f$ over all points $x,y$ in the domain; the \emph{local} behavior of gradient might be much more regular, indicating that a larger learning rate (and hence, faster convergence rate) might be permissible. In any case, it is beneficial to consider a modified gradient descent algorithm which, starting from a large initial learning rate, decreases the learning rate according to gradient information received so far, and stabilizes at at a rate depends on the local smoothness behavior and so no smaller than $1/L$.

  We consider the following modified gradient descent scheme:
\begin{figure}[H]
  \centering
  \begin{minipage}{.5\linewidth}
\begin{algorithm}[H]
   \caption{WNGrad -- Batch Setting}
   \label{alg:batch}
\begin{algorithmic}
   \STATE {\bfseries Input:} Tolerance $\epsilon > 0$ \\
    \STATE Initialize $x_1 \in \mathbb{R}^d, b_{1}>0, j \leftarrow 1$
    \REPEAT
  \STATE{$j  \leftarrow j+1$}
  \STATE $x_{j} \leftarrow x_{j-1} -  \frac{1}{b_{j-1}}\nabla f(x_{j-1})$
   \STATE $b_{j} \leftarrow  b_{j-1} + \frac{ \| \nabla f(x_{j}) \|^2}{b_{j-1}}$
    \UNTIL{$\| \nabla f(x_j) \|^2 \leq \epsilon$}  
\end{algorithmic}
\end{algorithm}

  \end{minipage}
\end{figure}

\begin{remark}
\textbf{Initializing $b_1$ and scale invariance.} Ideally, one could initialize $b_1$ in WNGrad by sampling $R \geq 2$ points $u_1, u_2, \dots, u_R$ close to the initialization $x_1$, and $\nabla f(u_1), \dots ,\nabla f(u_R)$, and take 
$$
b_1 = \max_{j \neq k} \frac{\| \nabla f(u_j) - \nabla f(u_k) \|}{ \| u_j - u_k \|} \leq  L
$$ 
If this is not possible, it is also reasonable to consider an initialization $b_1 = C \| \nabla f(x_1) \|$ with a constant $C\geq 0$.  With either choice, one observes that the resulting WNGrad algorithm is invariant to the scale of $f$: if $f$ is replaced by $\lambda f$, then the sequence of iterates $x_1, x_2, \dots$ remains unchanged.
\end{remark}

We show that the WNGrad algorithm has the following properties:
\begin{itemize}
\item After a reasonable number of initial iterations, either $\| \nabla f(x_k) \|^2 \leq  \epsilon$ or $b_k \geq L$ 
\item If at some point $b_k \geq L$, then the learning rate stabilizes: $b_j \leq C L$ for all $j \geq k$.
\end{itemize}
As a consequence, we have the following convergence result.

\begin{theorem}[Global convergence for smooth loss function]
\label{thm:converge}
Consider the WNGrad algorithm. Set $ b_1\geq  \| \nabla f(x_1) \|$. Suppose that $f \in C_L^1$, $x^{*}$ is the point satisfying $\nabla f(x^{*}) = 0.$ and that $f^{*} > -\infty$.  

Then we have the guarantee
$$
\min_{k=1:T} \| \nabla f(x_k) \|^2 \leq \epsilon
$$
\begin{enumerate}
\item[after] 
\subitem\textbf{Case 1} $\quad $  $T = \frac{2(f(x_1) - f^{*})(b_1 + 8  (f(x_1) - f^{*}))}{\epsilon}$ steps if $b_1 \geq  L,$ and \\
\subitem\textbf{Case 2}$\quad $ $T = 1+\frac{L^2(1-\delta)}{\epsilon} + \frac{16 \left((f(x_1) - f^{*}) +  (\frac{3}{16} + \frac{5}{8\delta})L\right)^2}{\epsilon}$ 
 steps if $b_1 = \delta L < L,  \quad \delta \in (0,1]$.
 \end{enumerate}

\end{theorem}

Comparing the convergence rate of batch gradient descent in Theorem \ref{thm:converge} and the classical convergence result in Lemma \ref{lem:classical}, we see that WNGrad adjusts the learning rate automatically with decreasing learning rate $1/b_j$ based on the gradient information received so far, and without knowledge of the constant L, and still achieves linear convergence at nearly the same rate as gradient descent in Lemma \ref{lem:classical} with constant learning rate $\eta \leq \frac{1}{L}$.  


We will use the following lemmas to prove Theorem \ref{thm:converge}. For more details, see Appendix A.1.

\begin{lemma}
\label{lem:increase}
Fix $\epsilon \in (0,1]$ and $L > 0$.  Consider the sequence
$$
b_1 > 0; \quad  \quad b_{j+1} = b_j +  \frac{\| \nabla f (x_{j+1}) \|^2}{b_j}
$$
after 
{$N = \max \left\{ 1, \lceil{ \frac{ L(L-b_1)}{\eta \epsilon} \rceil}+1 \right\}$}
iterations, either $\min_{k=1:N}  \| \nabla f(x_k) \|^2 \leq \epsilon$, or $b_{N} >  L$. 
\end{lemma}

\begin{lemma}
\label{lem:stabilize1}
Suppose that $f \in C_L^1$,  $f^{*} > -\infty$ and $b_1\geq \|\nabla f(x_1)\|$.  
Denote by $k_0$ the first index such that $b_{k_0} > L$.  Then for all $k \geq k_0$,
\begin{align}
b_{k} &\leq b_{k_0} +  8(f(x_{k_0})- f^{*})\nonumber 
\end{align}
and moreover, if $k_0 > 1,$
\begin{align}
 f(x_{k_0}) \leq f(x_1)  +\frac{L^2}{2b_1} .    \nonumber
\end{align}
\end{lemma}

Lemma \ref{lem:stabilize1} guarantees that the learning rate stabilizes once it reaches the (unknown) Lipschitz constant, up to an additive term.
To be complete, we can also bound $b_{k_0}$ as a function of $L$, then arrive at the main result of this section.

\begin{lemma}
\label{lem:stabilize3}
Suppose that $f \in C_L^1$ and that $f^{*} > -\infty$.  
Denote by $k_0$ the first index such that $b_{k_0} > L$.  Then
$$
b_{k_0} \leq 3 L + \frac{2L^2 }{b_1}.
$$
\end{lemma}
\section{WNGrad for Stochastic Gradient Descent}

We now shift from the setting of batch gradient descent to stochastic gradient descent.  The update rule to the learning rate in WNGrad extends without modification to this setting, but now that the gradient norms do not converge to zero but rather remain noisy, the WNGrad learning rate $\frac{1}{b_k}$ does not converge to a fixed size, but rather settles eventually on the rate of $\frac{G}{\sqrt{k}}$, where $G$ is a bound on the variance of the stochastic gradients. In order to tackle this issue and derive a convergence rate, we assume for the analysis that the loss function is convex but not necessarily smooth.

\begin{figure}[H]
  \centering
  \begin{minipage}{.5\linewidth}
\begin{algorithm}[H]
  \caption{WNGrad -- Stochastic Setting}
  \label{alg:stoch}
\begin{algorithmic}
  \STATE {\bfseries Input:} Tolerance $\epsilon > 0$
  \STATE Initialize $x_1\in \mathbb{R}^d$, $b_1 >0$, $j\leftarrow 1$
    \REPEAT
  \STATE{$j \leftarrow j+1$}
  \STATE $x_{j} \leftarrow x_{j-1} -  \frac{1}{b_{j-1}}g_{j-1}$
  \STATE $b_{j} \leftarrow b_{j-1} + \frac{ \| g_{j} \|^2}{b_{j-1}}$
    \UNTIL{$f(x_j) \leq \epsilon$} 
\end{algorithmic}
\end{algorithm}
  \end{minipage}
\end{figure}
Consider the general optimization problem
\begin{equation}
\min_x \quad f(x) \nonumber
\end{equation}
from stochastic gradient information.  Instead of observing full gradients $\nabla f(x_k)$, we observe stochastic gradients $g_k \in \mathbb{R}^d$ satisfying $\mathbb{E}(g_k) = \nabla f(x_k)$. Let $\overline{x}_k =\frac{1}{k}\sum_{i=1}^{k}x_i$.

\begin{theorem}
\label{thm:polarSGD}
Consider  WNGrad algorithm. Suppose $f(x)$ is convex.  Suppose, that, independent of $x_k$,
$$
\mathbb{E} \| g_k \|^2 \leq G^2
$$
 and that for all $k$
$$
 \| g_k \| \geq \gamma
$$
and $\mathbb{E} \| x_{k} - x^{*} \|^2 \leq D^2$.   
Then, with initialization  $b_1\geq\|g_1\|$,
$$
f(\overline{x}_k) - f^{*} \leq   \frac{G^2 (D^2 + 2)}{\gamma\sqrt{k}} + \frac{ b_1\| x_1 - x^{*} \|^2 }{2k}.
$$
\end{theorem}
\begin{remark}
Under the same assumptions, excluding the assumption that $\gamma^2 \leq \| g_k \|^2$, one obtains the same convergence rate using decreasing learning rate $\eta_k = \frac{c}{\sqrt{k}}$ for some constant $c$. 
\end{remark}
We will use the following lemma, which is easily proved by induction.

\begin{lemma}
\label{lem:sqroot}
Consider a positive constant $a > 0$ and a sequence of positive numbers $t_1, t_2, \dots$ and for each $k$,
$$
t_k + \frac{a}{t_k} \leq  t_{k+1}.
$$
Then, 
$$
 \sqrt{2ak} \leq t_k
$$
\end{lemma}

{\bf Proof of Theorem \ref{thm:polarSGD}:}
First, note that under the stated assumptions, $b_k$ satisfies Lemma \ref{lem:sqroot} for 
$a =  \gamma^2$.  Thus, with probability 1,
$$ 
b_k \geq \gamma \sqrt{k}
$$
Now,
$$ \| x_{k+1} - x^{*} \|^2 = \| x_{k} - x^{*} \|^2 + \frac{1}{b_k^2} \| g_k \|^2 - 2\frac{1}{b_k} \langle{ x_{k} - x^{*}, g_k \rangle},$$
so
$$
2 \langle{ x_{k} - x^{*}, g_k \rangle} =  {b_k} \| x_{k} - x^{*} \|^2 - {b_k} \| x_{k+1} - x^{*} \|^2  + \frac{1}{b_k} \| g_k \|^2  
$$
Thus,
\begin{align}
2 \sum_{\ell=1}^k \langle{ x_{\ell} - x^{*}, g_{\ell} \rangle} & \leq 2 \sum_{\ell=1}^k \langle{ x_{\ell} - x^{*}, g_{\ell} \rangle} + {b_k} \| x_{k+1} - x^{*} \|^2 \nonumber \\
&= \sum_{\ell=2}^k (b_{\ell} - b_{\ell-1})  \| x_{\ell} - x^{*} \|^2  +{b_1} \| x_1 - x^{*} \|^2   + \sum_{\ell=1}^k \frac{1}{b_{\ell}}\| g_{\ell} \|^2  \nonumber \\
&=  \sum_{\ell=2}^k \frac{  \| g_{\ell} \|^2}{b_{\ell-1}} \| x_{\ell} - x^{*} \|^2 + {b_1}\| x_1 - x^{*} \|^2+ \sum_{\ell=1}^k \frac{1}{b_{\ell}}\| g_{\ell} \|^2  \nonumber  \\
&\leq  \sum_{\ell=1}^{k-1} \frac{\| g_{\ell+1} \|^2}{\sqrt{\ell}  \gamma}  \| x_{\ell+1} - x^{*} \|^2  + {b_1}\| x_1 - x^{*} \|^2+ \sum_{\ell=1}^k \frac{1}{\sqrt{\ell} \gamma} \| g_{\ell} \|^2  \nonumber 
\end{align}
Now,  since $\langle{ x_k - x^{*},\nabla f(x_k) \rangle} = \mathbb{E} \langle{ x_k - x^{*}, g_k \rangle}$ and since $\mathbb{E} \| g_{\ell} \|^2 \leq G^2,$ conditioned on $g_{\ell-1}, \dots, g_1$, we apply the law of iterated expectation to obtain 
\begin{align}
2 \sum_{\ell=1}^k \langle{ x_{\ell} - x^{*}, g_{\ell} \rangle} 
&\leq \frac{1}{ \gamma} \left( \sum_{\ell=1}^{k-1} \frac{1}{\sqrt{\ell}} G^2 \mathbb{E}( \| x_{\ell+1} - x^{*} \|^2) \right)  + {b_1}\| x_1 - x^{*} \|^2 + \frac{1}{\gamma} \sum_{\ell=1}^k \frac{1 }{\sqrt{\ell}} G^2 \nonumber \\
&\leq \frac{2G^2 (D^2 + 1)}{\gamma}\sqrt{k-1} + {b_1}\| x_1 - x^{*} \|^2 + \frac{ 2}{\gamma}G^2 \sqrt{k}  \nonumber 
\end{align}
where in the final inequality, we use that $\sum_{\ell=1}^{k} \frac{1}{\sqrt{\ell}} \leq 2\sqrt{k}$.

From Jensens inequality, and recalling that by convexity $f(x_k) - f^{*} \leq \langle{ x_k - x^{*},\nabla f(x_k) \rangle} = \mathbb{E} \langle{ x_k - x^{*}, g_k \rangle} $, 
we conclude 
\begin{align}
f(\overline{x}_k) - f^{*}
&\leq\frac{1}{k} \sum_{\ell=1}^k  (f(x_{\ell}) - f^{*}) \nonumber \\
&\leq \frac{G^2 (D^2 + 1)\sqrt{k-1} + \gamma (b_1/2)\| x_1 - x^{*} \|^2 + G^2  \sqrt{k}}{\gamma k}\nonumber\\
&\leq \frac{G^2 (D^2 + 2)\sqrt{k} + \gamma (b_1/2)\| x_1 - x^{*} \|^2}{\gamma k}. \nonumber
\end{align}
\section{Numerical Experiments}
{ With guaranteed convergence of WNGrad in both batch and stochastic settings under appropriate conditions\footnote{We assume non-convex smooth loss function in batch setting and convex not necessarily smooth in stochastic setting}, we perform experiments in this section to show that WNGrad exhibits the same robustness for highly non-convex loss functions associated to deep learning problems. 

Consider a loss function $f$ whose gradient has Lipschitz constant $L$.  Then, the gradient of the rescaled loss function $\lambda f$ has Lipschitz constant $\lambda L$.  If we were to also rescale $b_1$ to $\lambda b_1$, then the dynamics $x_j \leftarrow x_{j-1}$ would remain unchanged due to scale invariance.  If instead we fix $b_1 = 1$ while letting $\lambda$ vary, we can test the robustness of WNGrad to different Lipschitz constants, and compare its robustness to  stochastic gradient descent (SGD, Algorithm \ref{alg:sgd} in Appendix).   To be precise, we consider the following variant of WNGrad, Algorithm \ref{alg:stoch_rescale}, and explore its performance as we vary $\lambda$.  Note that $\lambda$ in this algorithm is analogous to the constant learning rate $\eta$ in weight normalization and batch normalization as discussed in \eqref{wnupdate_sphere}.
\begin{figure}[H]
  \centering
  \begin{minipage}{.5\linewidth}
\begin{algorithm}[H]
   \caption{WNGrad - Scaled Loss Function}
   \label{alg:stoch_rescale}
\begin{algorithmic}
   \STATE {\bfseries Input:} Tolerance $\epsilon > 0$ \\
    \STATE Initialize $x_1 \in \mathbb{R}^d, b_{1}\leftarrow 1, j \leftarrow 1$
    \REPEAT
  \STATE{$j  \leftarrow j+1$}
     \STATE $b_{j} \leftarrow  b_{j-1} + \frac{ \lambda^2\| g_{j-1} \|^2}{b_{j-1}}$
  \STATE $x_{j} \leftarrow x_{j-1} -  \frac{\lambda g_{j-1}}{b_{j}}$
    \UNTIL{$\| \nabla f(x_j) \|^2 \leq \epsilon$}  
\end{algorithmic}
\end{algorithm}
\end{minipage}
\end{figure}
 WNGrad is mainly tested on two data sets: MNIST \cite{lecun1998gradient} and  CIFAR-10 \cite{krizhevsky2009learning}.
Table \ref{data-table} is the summary.  We use batch size 100  for both  MNIST and CIFAR-10.
 The experiments are done in PyTorch and  parameters are by default if no specification is provided. The data sets are preprocessed  with  normalization using mean and standard deviation of the entire train samples. Details in implementing WNGrad in a neural network are explained in Appendix A.3.
 
 We first test a wide range of the scale of the loss function\footnote{$\lambda \in\{10^{-0.25j+1.25}, j\in\{0,1,2,\cdots,19\} \} $} with two fully connected layers (without bias term) on MNIST (input dimension is $784$) 
 in a very simple setting excluding other factors that come into effect, such as regularization (weight decay), dropout, momentum, batch normalization, etc. 
In addition, we repeat 5 times for each experiment in order to avoid the initialization effect since random initialization of weight vectors is used in our experiments.

 The outcome of the experiments shown in Figure \ref{mnistFig11} verifies that WNGrad is very robust to the Lipschitz constant, while SGD is much more sensitive. \emph{This shows that the learning rate can be initialized at a high value if we consider $\lambda$ to be the learning rate.} When picking  $\lambda=0.562$ and $\lambda =0.056$, we have the train/test loss with respect to epoch shown in blue and dark-red curves respectively. With larger scale of Lipschitz constant ($\lambda=0.562$),  WNGrad does much better than SGD in both training and test loss. It is interesting to note that even with smaller scale of the Lipschitz constant $\lambda =0.056$, even thought SGD obtains the smaller training loss but does worse in generalization. On the contrary,  WNGrad gives better generalization (smaller test loss) despite of the larger train loss. Thus,  WNGrad to some extend is not only robust to the scale of Lipschitz constant but also generalizes well -- we aim to study this property of WNGrad in future work. 
\begin{figure}[H]
\centering
\includegraphics[width=.85\columnwidth]{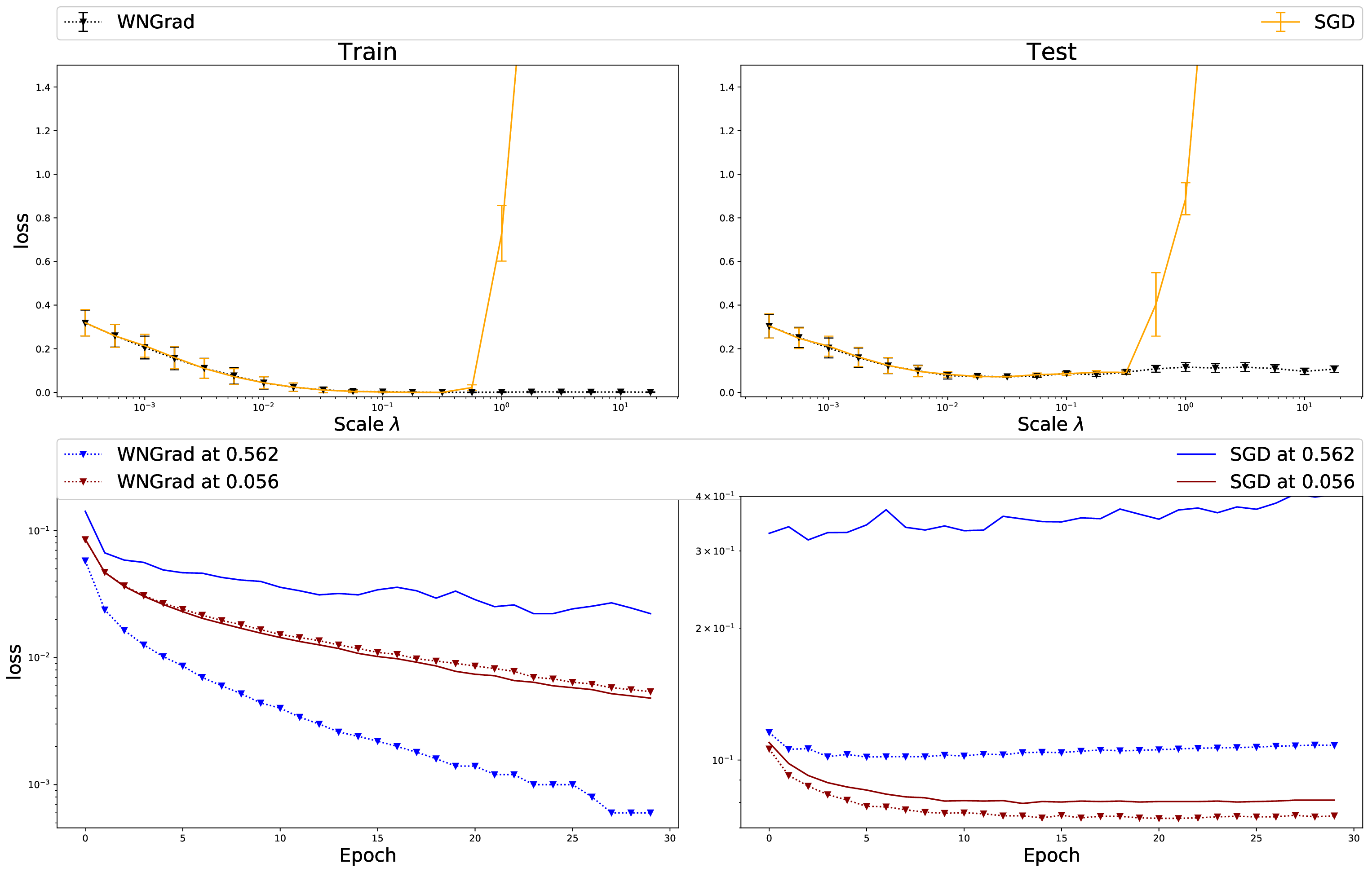}
\caption{Two fully connected layers on MNIST. The first row are plots of  mean Train/Test loss of 5 repeated experiments with respect to the scale of Lipschitz constant at epoch 30 and the second row are plots of  mean loss with respect to epoch at $\lambda=0.562$ (blue curves) and $\lambda=0.056$ (dark-red curves). The error bar of the plots in the first row means one standard deviation of five repeated experiments and no error bars shows in the second row for neatness. Better read on screen.}
\label{mnistFig11}
\begin{center}
\centerline{\includegraphics[width=.85\columnwidth]{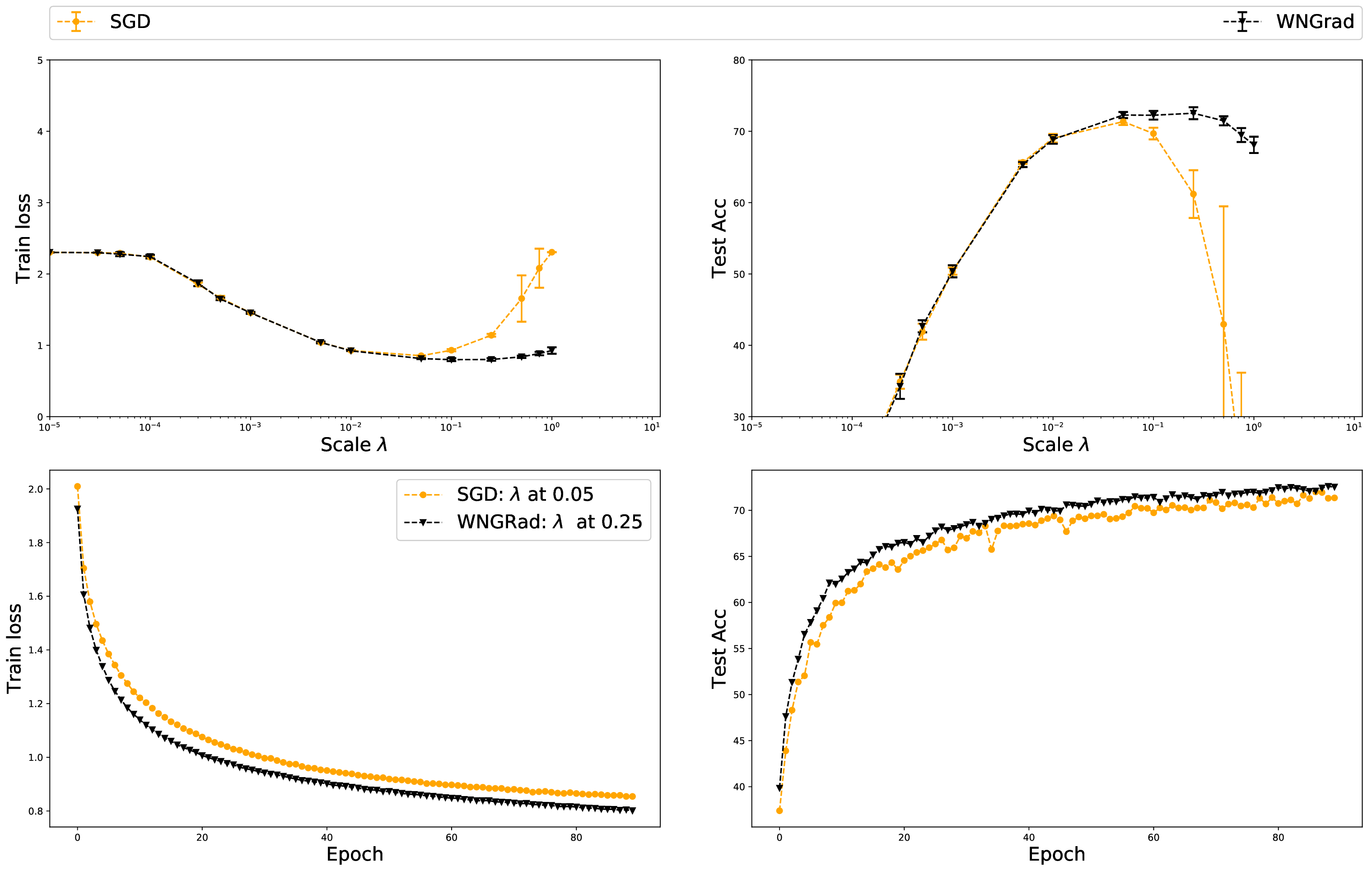}}
\caption{The simple convolutional network on CIFAR10. Top left (right) is the average training loss (test accuracy) of 5 repeated experiments with respect to the scale of Lipschitz constant at epoch 90. Bottom  left (right) is  the plot of mean train loss  (test accuracy)  with respect to epoch at the best  $\lambda$ found at this epoch. The error bar of the plots in the first row means one standard deviation of five repeated experiments and no error bars shown in the second row for neatness. Better read on screen.}
\label{cifar10Fig}
\end{center}
\end{figure}

Now we continue to compare the methods on a larger dataset, CIFAR10, with a wide range of scale $\lambda$ \footnote{
$ \lambda\in  \{ 1,\frac{3}{4}, \frac{1}{2}, \frac{1}{4}, \frac{1}{10},\frac{1}{20},\frac{1}{100}, \frac{1}{200}, \frac{1}{1000}, \frac{1}{2000},\frac{1}{10000}\}$. } from  $0.0001$ to $1$. We apply a simple convolution neural network (see Table \ref{2cnn} for details) with weight decay $10^{-4}$, of which the result shown in Figure \ref{cifar10Fig}.
In comparison with SGD, WNGrad is very robust to the scale $\lambda$ -- it performs better at  $ \lambda \in [0.01,1]$ and does as well as SGD when $ \lambda$ smaller than $0.01$. When at the best  $\lambda$ for each algorithm ($ \lambda=0.25$ WNGrad and $ \lambda=0.05$ SGD), WNGrad outperforms in training and testing along the way.
 

 A common practice to train deep and recurrent neural neural networks is to add momentum to stochastic gradient descent \cite{sutskever2013importance}. Recent adaptive moment estimation (Adam) \cite{kingma2014adam} seems to improve performance of models on a number of datasets. However, these methods are considerably sensitive to the scale of Lipschitz constant and require careful tuning in order to obtain the best result. Here we incorporate our algorithms with momentum  (WNGrad-Momentum) and adapt ``Adam" way (WN-Adam, Algorithm \ref{alg:adam}) in the hope to improve the robustness to the relationship between the learning rate and the Lipschitz constant. We use ResNet-18 training on CIFAR10 in Figure \ref{momentum}. Because of the batch normalization designed in  ResNet-18, we widen the range of $\lambda$ up to $10$. As we can see, WN-Adam (green curve) and WNGrad-Momentum  (black curve)  do seem to be more robust compared to Adam (red) and SGD-Momentum (orange). Particularly, WN-Adam is very robust even at $ \lambda =10$ and  still does fairly well in generalization. 
\begin{figure}[H]
  \centering
  \begin{minipage}{.5\linewidth}
\begin{algorithm}[H]
  \caption{WN-Adam}
  \label{alg:adam}
\begin{algorithmic}
  \STATE {\bfseries Input:} Tolerance $ \epsilon > 0$
  \STATE Initialize $x_1 \in \mathbb{R}^d,\hat{g}_{1}\leftarrow 0, b_{1}\leftarrow 1,j\leftarrow1 $
    \REPEAT
  \STATE{$j \leftarrow j+1$}
    \STATE $\hat{g}_{j} \leftarrow \beta_{1}\hat{g}_{j-1}+(1-\beta_{1})g_{j-1}, \beta_{1}=0.9$
     \STATE $b_{j}\leftarrow b_{j-1}+\lambda^2\frac{\|g_{j-1}\|^2}{b_{j-1}}$
    \STATE $x_{j} \leftarrow x_{j-1}-\frac{\lambda}{b_{j}}\frac{\hat{g}_{j}}{1-\beta_{1}^{j-1}}$
    \UNTIL{$f(x_j) \leq \epsilon$}  
\end{algorithmic}
\end{algorithm}
  \end{minipage}
  \begin{center}
\centerline{\includegraphics[width=0.8\columnwidth]{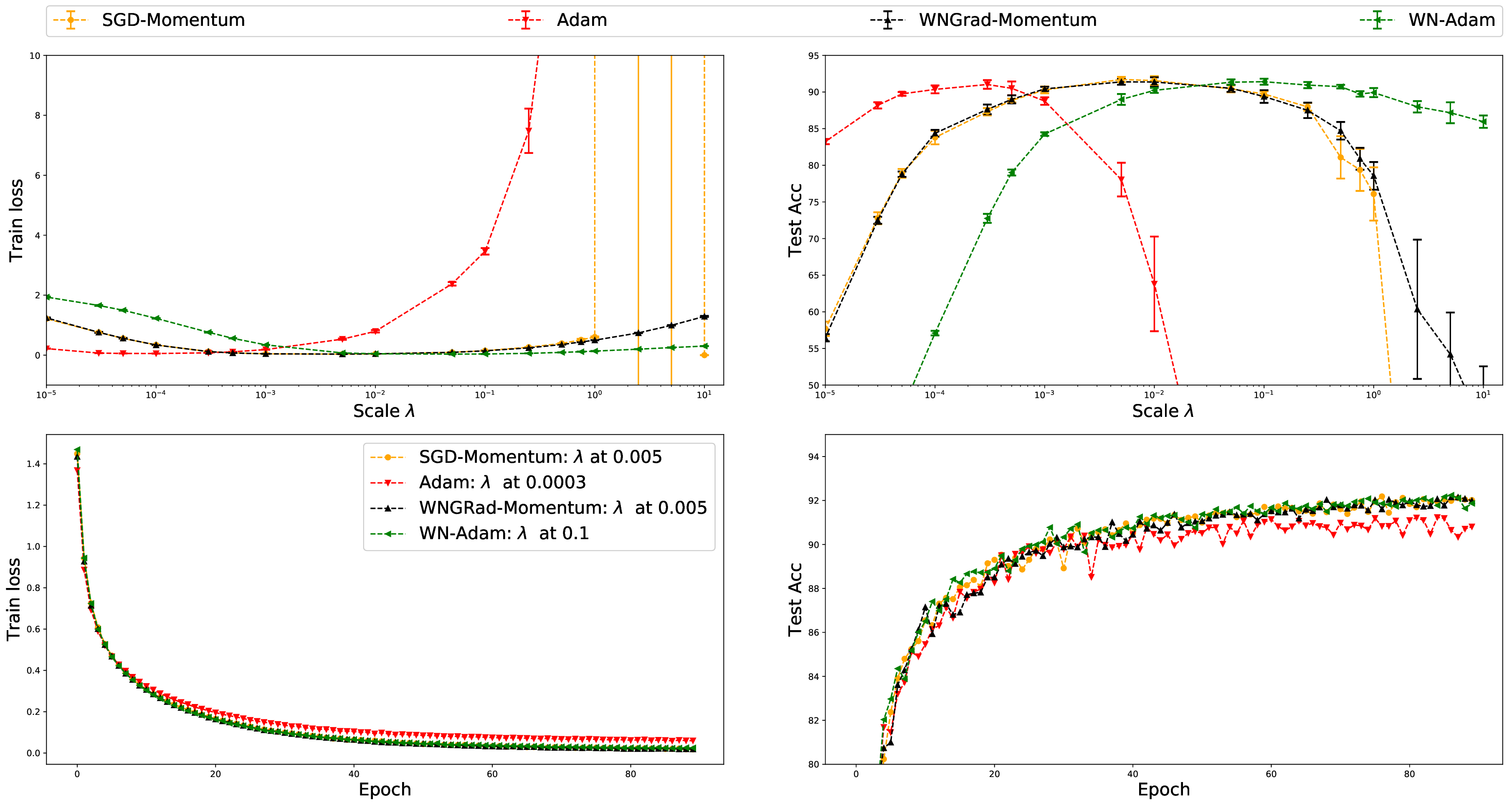}}
\caption{ResNet-18 on CIFAR10. Top plots are the snapshots of training at the 60th epoch. Reading instruction, see Figure \ref{cifar10Fig}.}
\label{momentum}
\end{center}
\end{figure}

\section{Conclusion}

We propose WNGrad, an method for dynamically updating the learning rate $1/b_k$ according to gradients received so far, which works in both batch and stochastic gradient methods and converges.  

In the batch gradient descent setting, we show that WNGrad converges to a weight vector $w_T$ satisfying $\| \nabla f(w_T) \|^2 \leq \epsilon$ in at most {$T = O(\frac{L+L^2}{\epsilon})$} iterations, if $f$ has $L$-Lipschitz smooth gradient.  This nearly matches the convergence rate for standard gradient descent with fixed learning rate $1/L$, but WNGrad does not need to know $L$ in advance. 

In the stochastic setting, the $b_k$ update in WNGrad has different behavior, growing like $O(\frac{\sqrt{k}}{G}),$ where $G$ is a bound on the variance of the stochastic gradients.  As a result, in the stochastic setting, we also show that WNGrad achieves the optimal $O(1/\sqrt{T})$ rate of convergence for convex loss functions, and moreover settles in expectation on the ``correct" rate, $b_k \sim \frac{\sqrt{k}}{G}$.  Thus, WNGrad works robustly in the stochastic setting, and finds a good learning rate. 

In numerical experiments, WNGrad competes favorably to plain stochastic gradient descent in terms of robustness to the relationship between the learning rate and the Lipschitz constant and generalization error in training neural networks on two standard data sets. And such robustness extends further to the algorithm that incorporates momentum (WN-Adam and WNGrad-Momentum).


\section*{Acknowledgments}
We thank Arthur Szlam and Mark Tygert for constructive suggestions. Also, we appreciate the help (with the experiments) from Sam Gross, Shubho Sengupta, Teng Li, Ailing Zhang, Zeming Lin, and Timothee Lacroix. 
\bibliography{icml2018}

\begin{thebibliography}{23}
\providecommand{\natexlab}[1]{#1}
\providecommand{\url}[1]{\texttt{#1}}
\expandafter\ifx\csname urlstyle\endcsname\relax
  \providecommand{\doi}[1]{doi: #1}\else
  \providecommand{\doi}{doi: \begingroup \urlstyle{rm}\Url}\fi

\bibitem[Alexandre and Francis(2017)]{defossez2017adabatch}
D.~Alexandre and B.~Francis.
\newblock Adabatch: Efficient gradient aggregation rules for sequential and
  parallel stochastic gradient methods.
\newblock \emph{arXiv preprint arXiv:1711.01761}, 2017.

\bibitem[Bach and Moulines(2011)]{bm11}
F.~Bach and E.~Moulines.
\newblock Non-asymptotic analysis of stochastic approximation algorithms for
  machine learning.
\newblock In \emph{Advances in Neural Information Processing Systems}, 2011.

\bibitem[Bottou et~al.(2016)Bottou, Curtis, and Nocedal]{bcn16}
L.~Bottou, F.~Curtis, and J.~Nocedal.
\newblock Optimization methods for large-scale machine learning.
\newblock In \emph{arXiv preprint arXiv:1606.04838}, 2016.

\bibitem[Dimitri(1999)]{bertsekas1999nonlinear}
B.~Dimitri.
\newblock \emph{Nonlinear programming}.
\newblock Athena scientific Belmont, 1999.

\bibitem[Duchi et~al.(2011)Duchi, Hazan, and Singer]{duchi2011adaptive}
J.~Duchi, E.~Hazan, and Y.~Singer.
\newblock Adaptive subgradient methods for online learning and stochastic
  optimization.
\newblock \emph{Journal of Machine Learning Research}, 12\penalty0
  (Jul):\penalty0 2121--2159, 2011.

\bibitem[Goyal et~al.(2017)Goyal, Doll{\'{a}}r, Girshick, Noordhuis,
  Wesolowski, Kyrola, Tulloch, Jia, and He]{GoyalDGNWKTJH17}
P.~Goyal, P.~Doll{\'{a}}r, R.~B. Girshick, P.~Noordhuis, L.~Wesolowski,
  A.~Kyrola, A.~Tulloch, Y.~Jia, and K.~He.
\newblock Accurate, large minibatch {SGD:} training imagenet in 1 hour.
\newblock \emph{CoRR}, abs/1706.02677, 2017.
\newblock URL \url{http://arxiv.org/abs/1706.02677}.

\bibitem[Ioffe and Szegedy(2015)]{ioffe2015batch}
S.~Ioffe and C.~Szegedy.
\newblock Batch normalization: Accelerating deep network training by reducing
  internal covariate shift.
\newblock In \emph{International conference on machine learning}, pages
  448--456, 2015.

\bibitem[Kingma and Ba(2014)]{kingma2014adam}
D.~Kingma and J.~Ba.
\newblock Adam: A method for stochastic optimization.
\newblock \emph{CoRR}, abs/1412.6980, 2014.
\newblock URL
  \url{http://dblp.uni-trier.de/db/journals/corr/corr1412.html#KingmaB14}.

\bibitem[Krizhevsky(2009)]{krizhevsky2009learning}
A.~Krizhevsky.
\newblock Learning multiple layers of features from tiny images.
\newblock 2009.

\bibitem[LeCun et~al.(1998{\natexlab{a}})LeCun, Bottou, Bengio, and
  Haffner]{lecun1998gradient}
Y.~LeCun, L.~Bottou, Y.~Bengio, and P.~Haffner.
\newblock Gradient-based learning applied to document recognition.
\newblock \emph{Proceedings of the IEEE}, 86\penalty0 (11):\penalty0
  2278--2324, 1998{\natexlab{a}}.

\bibitem[LeCun et~al.(1998{\natexlab{b}})LeCun, Bottou, Orr, and
  M{\"u}ller]{lecun1998efficient}
Y.~LeCun, L.~Bottou, G.~Orr, and K.~M{\"u}ller.
\newblock Efficient backprop.
\newblock In \emph{Neural networks: Tricks of the trade}, pages 9--50.
  Springer, 1998{\natexlab{b}}.

\bibitem[Needell et~al.(2014)Needell, Ward, and Srebro]{nsw14}
D.~Needell, R.~Ward, and N.~Srebro.
\newblock Stochastic gradient descent, weighted sampling, and the randomized
  kaczmarz algorithm.
\newblock In \emph{Advances in Neural Information Processing Systems}, 2014.

\bibitem[Nesterov(1998)]{nesterov1998introductory}
Y.~Nesterov.
\newblock Introductory lectures on convex programming volume i: Basic course.
\newblock 1998.

\bibitem[Robbins and Monro(1951)]{rm51}
H.~Robbins and S.~Monro.
\newblock A stochastic approximation method.
\newblock In \emph{The Annals of Mathematical Statistics}, volume~22, pages
  400--407, 1951.

\bibitem[Salimans and Kingma(2016)]{salimans2016weight}
T.~Salimans and D.~Kingma.
\newblock Weight normalization: A simple reparameterization to accelerate
  training of deep neural networks.
\newblock In \emph{Advances in Neural Information Processing Systems}, pages
  901--909, 2016.

\bibitem[Shamir and Zhang(2013)]{shamir2013stochastic}
O.~Shamir and T.~Zhang.
\newblock Stochastic gradient descent for non-smooth optimization: Convergence
  results and optimal averaging schemes.
\newblock In \emph{International Conference on Machine Learning}, pages 71--79,
  2013.

\bibitem[Srivastava and Swersky()]{hinton2012neural}
G.~Hinton~N. Srivastava and K.~Swersky.
\newblock Neural networks for machine learning-lecture 6a-overview of
  mini-batch gradient descent.

\bibitem[Sutskever et~al.(2013)Sutskever, Martens, Dahl, and
  Hinton]{sutskever2013importance}
I.~Sutskever, J.~Martens, G.~Dahl, and G.~Hinton.
\newblock On the importance of initialization and momentum in deep learning.
\newblock In \emph{International conference on machine learning}, pages
  1139--1147, 2013.

\bibitem[Ward et~al.(2019)Ward, Wu, and Bottou]{ward2019adagrad}
R.~Ward, X.~Wu, and L.~Bottou.
\newblock Adagrad stepsizes: Sharp convergence over nonconvex landscapes.
\newblock In \emph{International Conference on Machine Learning}, pages
  6677--6686, 2019.

\bibitem[Wilson et~al.(2017)Wilson, Roelofs, Stern, Srebro, and
  Recht]{wilson2017marginal}
A.~Wilson, R.~Roelofs, M.~Stern, N.~Srebro, and B.~Recht.
\newblock The marginal value of adaptive gradient methods in machine learning.
\newblock In \emph{Advances in Neural Information Processing Systems}, pages
  4151--4161, 2017.

\bibitem[Wright and Nocedal(2006)]{Nocedal2006}
S.~Wright and J.~Nocedal.
\newblock \emph{Numerical Optimization}.
\newblock Springer New York, New York, NY, 2006.
\newblock ISBN 978-0-387-40065-5.
\newblock \doi{10.1007/978-0-387-40065-5_3}.
\newblock URL \url{https://doi.org/10.1007/978-0-387-40065-5_3}.

\bibitem[Zeiler(2012)]{zeiler2012adadelta}
M.~Zeiler.
\newblock Adadelta: an adaptive learning rate method.
\newblock In \emph{arXiv preprint arXiv:1212.5701}, 2012.

\bibitem[Zhao and Zhang(2015)]{zhao2015stochastic}
P.~Zhao and T.~Zhang.
\newblock Stochastic optimization with importance sampling for regularized loss
  minimization.
\newblock In \emph{International Conference on Machine Learning (ICML)}, pages
  1--9, 2015.

\end{thebibliography}


\appendix
\section{Appendix}
\subsection{Proof ingredients}
\begin{lemma}[Descent Lemma]
\label{lem:descend}
Let $f \in C_L^1$, i.e., $\forall x, y \in \mathbb{R}^d$,
$
\| \nabla f(x) - \nabla f(y) \| \leq L \| x - y \|.
$.  Then,
$$
f(x) \leq f(y) + \langle{\nabla f(y), x-y \rangle} + \frac{L}{2} \| x - y \|^2
$$
\end{lemma}
\subsubsection{Proof of Lemma \ref{lem:increase}}
{\bf Proof:}
If $b_1 \geq  L$, we are done.  So suppose $b_1 \leq b_{N} <  L$.   Thus,
\begin{align*}
 L > b_{N} &= b_1 +  \sum_{k=1}^{N-1} \frac{ \| \nabla f(x_{k+1}) \|^2}{b_k} \nonumber \\
&>  b_1 +  \sum_{k=1}^{N-1} \frac{\| \nabla f(x_{k+1}) \|^2}{  L} \nonumber .
\end{align*}
So $\sum_{k=2}^{N} \| \nabla f(x_k) \|^2 \leq   L( L - b_1),$ and hence 
{  \begin{align}
\min_{k=1:N}\| \nabla f(x_k) \|^2 &\leq \frac{1}{N-1} \sum_{k=2}^{N} \| \nabla f(x_k) \|^2 \nonumber \\
&\leq \frac{L( L  - b_1)}{ (N-1)} \leq \epsilon. \nonumber
\end{align}}

\subsubsection{Proof of Lemma \ref{lem:stabilize1}}
Suppose $k_0$  is the first index such that $b_{k_0} >  L$.  Then $b_{j} >  L$ for all $j \geq k_0$, and by Lemma \ref{lem:descend}, for $ j\geq k_0,$
\begin{align}
f(x_{j+1}) &\leq f(x_{j}) - \frac{1}{b_{j}}(1 - \frac{L}{2b_{j}}) \| \nabla f(x_{j}) \|^2 \nonumber \\
&\leq f(x_{j}) - \frac{1}{2 b_{j}} \| \nabla f(x_{j}) \|^2 \nonumber \\
&\leq f(x_{j}) - \frac{1}{2 b_{j+1}} \| \nabla f(x_{j}) \|^2 \nonumber \\
&\leq f(x_{k_0}) - \sum_{\ell=1}^{j} \frac{1}{2 b_{k_0+\ell-1}} \| \nabla f(x_{k_0+\ell-1}) \|^2 \nonumber 
\end{align}
Taking $j \rightarrow \infty$,
$$
 \sum_{\ell=1}^{\infty} \frac{ \| \nabla f(x_{k_0+\ell-1}) \|^2}{ b_{k_0+\ell-1}} \leq {2(f(x_{k_0})- f^{*})}.
$$
Now, if $k_0 > 1,$ then
\begin{align}
\label{boundb}
 b_{k_0+j} -  b_{k_0}&=  \sum_{\ell = 1}^j \frac{ \| \nabla f(x_{k_0+\ell}) \|^2}{b_{k_0+\ell-1}} \nonumber \\
&\leq  2 \sum_{\ell = 1}^j \frac{ \|\nabla f(x_{k_0+\ell-1}) - \nabla f(x_{k_0+\ell}) \|^2 + \|\nabla f(x_{k_0+\ell-1}) \|^2}{b_{k_0+\ell-1}}  \nonumber \\
&\leq  2 \sum_{\ell = 1}^j \frac{ L^2 \| x_{k_0+\ell-1} - x_{k_0+\ell} \|^2+\|\nabla f(x_{k_0+\ell-1}) \|^2}{b_{k_0+\ell-1}}  \nonumber \\
&= 2  \sum_{\ell = 1}^j \frac{ L^2  \| \nabla f(x_{k_0+\ell-1}) \|^2}{b_{k_0+\ell-1}^3} + 2 \sum_{\ell = 1}^j \frac{ \|\nabla f(x_{k_0+\ell-1}) \|^2}{b_{k_0+\ell-1}} \nonumber \\
&\leq 4  \sum_{\ell = 1}^j \frac{ \|\nabla f(x_{k_0+\ell-1}) \|^2}{b_{k_0+\ell-1}} \nonumber \\
&\leq   8 (f(x_{k_0})- f^{*}) 
\end{align}
since $b_{k_0}\geq  L$. Finally, since $b_j \leq  L$ for $j = 1=1,2,\dots, k_0-1$, we can bound $f(x_{k_0})-f^{*}$.  By Lemma \ref{lem:descend},
\begin{align}
f(x_{k_0}) &\leq f(x_1) + \frac{L }{2} \sum_{j=1}^{{k_0}-1}  \frac{\| \nabla f(x_j) \|^2}{b_j^2} \nonumber \\
&\leq f(x_1) + \frac{L  \| \nabla f(x_1) \|^2}{2b_1^2} +  \frac{L }{2} \sum_{j=1}^{{k_0}-2}  \frac{\| \nabla f(x_{j+1}) \|^2}{b_{j}^2} \nonumber \\
&\leq f(x_1) + \frac{L  \| \nabla f(x_1) \|^2}{2b_1^2} +  \frac{L}{2b_1} (b_{k_0-1} - b_1)  \nonumber \\
&\leq f(x_1)  +  \frac{Lb_{k_0-1}}{2b_1}   
\end{align}
since $b_1\geq  \|\nabla f(x_1)\|$ and $b_{k_0-1}\leq  L $.
\subsubsection{Proof of Lemma \ref{lem:stabilize3}}
We use shorthand $\nabla f_k = \nabla f(x_k)$. Let $k_0\geq 1$ be the first index such that $b_{k_0}\geq  L$. Then,
\begin{align}
b_{k_0} = b_{k_0-1} +  \frac{ \| \nabla f_{k_0} \|^2}{ b_{k_0-1}}
&\leq b_{k_0-1} + 2 \frac{ \| \nabla f_{k_0} - \nabla f_{k_0-1} \|^2 + \| \nabla f_{k_0-1} \|^2 }{ b_{k_0-1}} \nonumber \\
&\leq b_{k_0-1} + 2 \left( \frac{ L^2 \| x_{k_0} - x_{k_0-1} \|^2}{ b_{k_0-1}} + \frac{\| \nabla f_{k_0-1} \|^2 }{ b_{k_0-1}} \right) \nonumber \\
&= b_{k_0-1} + 2 \left( \frac{ L^2  \| \nabla f_{k_0-1} \|^2}{ b_{k_0-1}^3 } + \frac{\| \nabla f_{k_0-1} \|^2 }{ b_{k_0-1}} \right) \nonumber \\
&\leq b_{k_0-1} + \frac{ 2L^2  \| \nabla f_{k_0-1} \|^2}{ b_{k_0-1}^2 b_{k_0-2} } + \frac{2 \| \nabla f_{k_0-1} \|^2 }{ b_{k_0-1}} \nonumber \\
&= b_{k_0-1} + \frac{ 2L^2  (b_{k_0-1} - b_{k_0-2})}{ b_{k_0-1}^2 } + \frac{2 \| \nabla f_{k_0-1} \|^2 }{ b_{k_0-1}} \nonumber \\
&\leq b_{k_0-1} + \frac{ 2L^2 }{b_{k_0-1}}  + \frac{2 \| \nabla f_{k_0-1} \|^2 }{ b_{k_0-2}} \nonumber \\
&\leq 3  L + \frac{2L^2 }{b_1}
\end{align}

\subsection{Proof of Theorem \ref{thm:converge}}
By Lemma \ref{lem:increase}, if $\min_{k=1:N} \| \nabla f(x_k) \|^2 \leq \epsilon$ is not satisfied after 
$N = \lceil{ \frac{L(  L - b_1)}{ \epsilon} \rceil} \leq \frac{(1-\delta)L^2}{\epsilon}$ steps, then there is a first index $k_0 \leq N$ such that $b_{k_0} >  L$.  By Lemma \ref{lem:stabilize1},  for all $k \geq k_0$,
$$
b_k \leq b_{k_0} + 8  (f(x_{k_0}) - f^{*}),
$$ 
so set 
$$
P = b_{k_0} + 8  (f(x_{k_0}) - f^{*}).
$$
If $k_0 = 1,$ then it follows that
\begin{align}
f(x_{M}) \leq f(x_1) - \frac{\sum_{k=1}^{M} \| \nabla f(x_{k}) \|^2}{2(b_{1} + 8  (f(x_{1}) - f^{*}))} 
\end{align}
and thus the stated result holds straightforwardly.

Otherwise, if $k_0 > 1,$ then, by Lemma \ref{lem:descend}, for any $M \geq 1$,
\begin{align}
f(x_{k_0+M}) &\leq f(x_{k_0+M-1}) - \frac{1}{2b_{k_0+M-1}} \| \nabla f(x_{k_0+M-1}) \|^2 \nonumber \\
 &\leq f(x_{k_0+M-1}) - \frac{1}{2P} \| \nabla f(x_{k_0+M-1}) \|^2 \nonumber \\
 &\leq f(x_{k_0}) - \frac{1}{2P} \sum_{k=1}^{M} \| \nabla f(x_{k_0+k-1}) \|^2. \nonumber 
\end{align}
By Lemma \ref{lem:stabilize3}, since $b_1 \geq \delta  L$, we have 
$$
b_{k_0} \leq (3 + \frac{2}{\delta})  L
$$
By Lemma \ref{lem:increase}, 
$$
f(x_{k_0}) - f^{*} \leq f(x_1) - f^{*} + \frac{L }{2\delta}, 
$$
Thus, 
\begin{align}
\min_{k=1:M}  \| \nabla f(x_{k_0+k-1}) \|^2 
&\leq \frac{1}{M} \sum_{k=1}^{M} \| \nabla f(x_{k_0+k-1}) \|^2 \nonumber \\
&\leq  \frac{2P(f(x_{k_0}) - f^{*})}{ M} \nonumber \\
 &{=  \frac{2(b_{k_0} + 8  (f(x_{k_0}) - f^{*}))(f(x_{k_0}) - f^{*})}{ M} \nonumber }\\
&\leq \frac{2b_{k_0}(f(x_{k_0}) - f^{*})}{ M} +\frac{16(f(x_{k_0}) - f^{*})^2}{M}\nonumber\\
&\leq \frac{2(3+\frac{2}{\delta})L(f(x_{k_0}) - f^{*})}{ M} +\frac{16(f(x_{k_0}) - f^{*})^2}{M}\nonumber\\
\end{align}
Thus, once
$$
M \geq \frac{16 (f(x_1) - f^{*} +  (\frac{3}{16} + \frac{5}{8\delta}) L)^2}{\epsilon},
$$
we are assured that
$$
\min_{k=1:N+M}  \| \nabla f(x_{k}) \|^2 \leq \epsilon
$$
where $N \leq \frac{L^2(1-\delta)}{\epsilon} $.


\subsection{Implementing the Algorithm in A Neural Network}
In this section, we give the details for implementing our algorithm in a neural network. In the standard neural network architecture, the computation of each neuron consists of an elementwise nonlinearity  of a linear transform of input features or output of previous layer:
\begin{equation}
y = \phi (\langle{w,x\rangle}+b)  \label{eqa:neuron},
\end{equation}
where $w$ is the $d$-dimensional weight vector, $b$ is a scalar bias term, $x$,$y$ are respectively a $d$-dimensional vector of input features (or output of previous layer) and the output of current neuron, $\phi(\cdot)$ denotes an elementwise nonlinearity. When using backpropogration \cite{lecun1998efficient} the stochastic gradient of $g$ in Algorithms \ref{alg:stoch}, \ref{alg:stoch_rescale} and \ref{alg:adam}  represent the gradient of the current neuron (see Figure \ref{fig:M1}).  Thus, when implementing our algorithm in PyTorch, WNGrad is one learning rate associated to one neuron, while SGD has one learning rate for all neurons.

\def\layersep{1.5cm}
\begin{figure}[H]
\centering
\begin{tikzpicture}[
   shorten >=1pt,->,
   draw = black!50,
    node distance=\layersep,
    every pin edge/.style={<-,shorten <=1pt},
    neuron/.style={circle,fill=black!25,minimum size=17pt,inner sep=0pt},
    input neuron/.style={neuron, fill=black!50},
    output neuron/.style={neuron, fill=black!50},
    hidden neuron/.style={neuron, fill=black!50},
    annot/.style={text width=4em, text centered}
]

    \foreach \name / \y in {1,...,4}
        \node[input neuron, pin=left:Dim \y] (I-\name) at (0,-\y) {};

    \newcommand\Nhidden{2}

    \foreach \N in {1,...,\Nhidden} {
       \foreach \y in {1,...,5} {
          \path[yshift=0.5cm]
              node[hidden neuron] (H\N-\y) at (\N*\layersep,-\y cm) {};
           }
   \node[annot,above of=H\N-1, node distance=1cm] (hl\N) {Hidden layer \N};
    }

    \node[output neuron,pin={[pin edge={<-}]right: loss}, right of=H\Nhidden-3] (O) {};

    \foreach \source in {1,...,4}
        \foreach \dest in {1,...,5}
            \path (H1-\dest)  edge  (I-\source);

    \foreach [remember=\N as \lastN (initially 1)] \N in {2,...,\Nhidden}
       \foreach \source in {1,...,5}
           \foreach \dest in {1,...,1}
               \path  (H\N-\dest) edge [green, thick, domain=-2:2] (H\lastN-\source) ;
    \foreach [remember=\N as \lastN (initially 1)] \N in {2,...,\Nhidden}
       \foreach \source in {1,...,5}
           \foreach \dest in {2,...,5}
               \path (H\N-\dest) edge   (H\lastN-\source);

    \foreach \source in {1,...,5}
        \path   (O)  edge (H\Nhidden-\source);


    \node[annot,left of=hl1] {Input layer};
    \node[annot,right of=hl\Nhidden] {Output layer};
\end{tikzpicture}
\caption{ An example of backproporgation of two hidden layers. Green edges represent the stochastic gradient  $g$ in Algorithm \ref{alg:stoch_rescale} and \ref{alg:adam}.  } \label{fig:M1}
\end{figure}

\subsection{Tables and Algorithms}
\begin{center}
 \begin{minipage}{.5\linewidth}
\begin{algorithm}[H]
   \caption{SGD }
   \label{alg:sgd}
\begin{algorithmic}
   \STATE {\bfseries Input:} Tolerance $\epsilon > 0$\\
    \STATE Initialize $x_1 \in \mathbb{R}^d, b_{1}\leftarrow 1, j \leftarrow 1$
    \REPEAT
  \STATE{$j  \leftarrow j+1$}
  \STATE $x_{j} \leftarrow x_{j-1} -   {\lambda g_{j-1}}$
    \UNTIL{$\| \nabla f(x_j) \|^2 \leq \epsilon$}  
\end{algorithmic}
\end{algorithm}
\end{minipage}
\end{center}

\begin{table}[H]
\caption{Statistics of data sets. DIM is the dimension of a sample}
\label{data-table}
\begin{center}
\begin{small}
\begin{sc}
\begin{tabular}{lcccr}
\toprule
Dataset &  Train  & Test & Classes & Dim\\
\midrule
MNIST    &   60,000& 10,000& 10& 28$\times$28 \\
CIFAR-10    & 50,000 & 10,000 &10 & 32$\times$32\\
\bottomrule
\end{tabular}
\end{sc}
\end{small}
\end{center}

\caption{architecture for five-layer convolution neural network }
\label{2cnn}
\begin{center}
\begin{small}
\begin{sc}
\begin{tabular}{cccr}
\toprule
Layer type &  Channels & Out Dimension\\
$5\times 5$ conv relu & 6 & 28\\
$2\times 2$ max pool, str.2 & 6 & 14\\
$5\times 5$ conv relu & 16 & 10\\
$2\times 2$ max pool, str.2 & 6 & 5\\
FC  relu& N/A & 120\\
FC  relu& N/A & 84\\
FC  relu& N/A & 10\\
\bottomrule
\end{tabular}
\end{sc}
\end{small}
\end{center}
\end{table}


\end{document}